\documentclass[runningheads,a4paper]{llncs}

\usepackage{amssymb}
\usepackage{graphicx}
\newcommand\inv[1]{#1\raisebox{1.15ex}{$\scriptscriptstyle-\!1$}}

\usepackage{times}
\usepackage{latexsym}
\usepackage{times}
\usepackage{url}
\usepackage{latexsym}
\usepackage{graphicx}
\usepackage{multirow}
\usepackage{booktabs}
\usepackage{pgfplots}
\usepackage{lscape}
\usepackage{tikz}
\usetikzlibrary{arrows,positioning,shapes.geometric}
\usepackage{amsmath}
\usepackage{algorithm}
\usepackage[noend]{algpseudocode}
\usepackage{array,multirow}
\usepackage{algpseudocode}
\usepackage{pifont}
\usepackage{graphicx}

\usepackage{hyperref}

\providecommand{\shortcite}[1]{\cite{#1}}

\authorrunning{Akhtar et al.}

\begin{document}

\mainmatter  

\title{An Unsupervised Approach for Mapping between Vector Spaces}

\author{\textbf{Syed Sarfaraz Akhtar* \qquad Arihant Gupta*\qquad Avijit Vajpayee*\qquad \\ Arjit Srivastava\qquad Madan Gopal Jhanwar\qquad Manish Shrivastava}\\  \{syed.akhtar, arihant.gupta, arjit.srivastava, madangopal.jhanwar\}@research.iiit.ac.in,\\ avijit@inshorts.com,\\ manish.shrivastava@iiit.ac.in}

\institute{Language Technologies Research Center(LTRC) \\
Kohli Center On Intelligent Systems (KCIS) \\
International Institute of Information Technology Hyderabad  (IIITH), 500032}


%
%
%


%
%

\toctitle{Lecture Notes in Computer Science}
\tocauthor{Authors' Instructions}
\maketitle

\begin{abstract}
We present a language independent, unsupervised approach for transforming word embeddings from source language to target language using a transformation matrix. Our model handles the problem of data scarcity which is faced by many languages in the world and yields improved word embeddings for words in the target language by relying on transformed embeddings of words of the source language. We initially evaluate our approach via word similarity tasks on a similar language pair - Hindi as source and Urdu as the target language, while we also evaluate our method on French and German as target languages and English as source language. Our approach improves the current state of the art results - by 13\% for French and 19\% for German. For Urdu, we saw an increment of 16\% over our initial baseline score. We further explore the prospects of our approach by applying it on multiple models of the same language and transferring words between the two models, thus solving the problem of missing words in a model. We evaluate this on word similarity and word analogy tasks.
\end{abstract}

\let\thefootnote\relax\footnotetext{* These authors contributed equally to this work.}
\section{Introduction}

Word representations are being widely used to solve problems of various areas of natural language processing. These include but are not limited to dependency parsing  ~\cite{BSL:14}, named entity recognition ~\cite{MIL:04} and parsing ~\cite{SOCH:13}.

In this paper, we try to exploit similarities in linguistically similar languages by transforming word embeddings of source language - which is highly resource rich and hence well trained, to a corresponding model of target language - which is relatively resource deficient. This technique is similar to that used by Mikolov et. al. \shortcite{MSG3:13c} for machine translation. We further extend our approach by applying it to different models trained for one particular language. This enables us to incorporate the best of various models.

Given a source of structured connections between words, Faruqui et.al \shortcite{MF:15} proposed a technique to combine embeddings learned from distributional semantics of unstructured text, known as ``retrofitting". Speer and Chin \shortcite{RS:16} extended this technique to produce state of the art word embeddings for English. The method proposed by them resulted in a 16\% increase on the Stanford English Rare-Word (RW) dataset~\cite{MAC:13}.

In this paper, we propose a method to transform words from one vector space to another. We use this technique to transform word embeddings between languages and also within the same language. Some languages are richer in resources than others. For example, Hindi is richer in resources than Urdu. We also evaluate our approach on dissimilar language pairs like English - French and English - German. Our aim is to use resource rich techniques like ConceptNet Ensemble \cite{RS:16} for languages poor in resources, for which we need to learn a mapping between their vector spaces.

Using this technique, we are also able to get embeddings of words which are unknown for one embedding space by importing the transformed embedding of the same word from another model of the same language. This method proved to be faster than training a combined embedding space from scratch, while giving high quality word embeddings.

The basis of our approach lies in having a sufficient number of frequent word pairs in both source and target languages to successfully train our transformation matrix. Each word pair is of form $<$word from source language, translated word of target language$>$. In this paper, we present a method for transforming word representations which, for training, take word representations of these word pairs to create a single transformation matrix, which when applied on any word representation of source word, will give us word representation of corresponding word in target language. We emphasize on highly frequent words, because we believe that highly frequent words have better trained word embeddings and thus result in better results for our approach.

We rely on a bi-lingual dictionary of the language pair for training and evaluating our transformation matrix. For training our matrix, we generate a bi-lingual dictionary from parallel corpus of source and target language in an unsupervised way. But for evaluation, the bi-lingual dictionary was missing most of the word pairs present in our test dataset because the parallel corpus for all the language pairs that we worked on in this paper were not large enough. Hence, only for evaluating our transformation matrix, we manually created a bilingual dictionary from the test sets - which was not used for training our transformation matrix because transformation matrix tends to completely remember the word representations it was generated from.

We show that our method performs well on both French and German with state of the art results on both languages. Since there is no prior work done on Urdu, we test our approach against baseline scores of SkipGram embeddings. Our approach showed improvement of 16\% over baseline scores. We further test our approach on multiple models of English. Our evaluations show that this is a fast and efficient approach to transform word embeddings from one vector space to another.

\section{Datasets}

For some experiments, we are using word embeddings trained on Google News corpus~\cite{MSG3:13c}. For all the models trained in this paper, we have used the Skip-gram~\cite{MSG:13a} algorithm. The dimensionality has been fixed at 300 with a minimum count of 5 along with negative sampling. 

As training set for English, we use the Wikipedia data~\cite{SW:10} (SG/en-SG). Soricut and Och \shortcite{SO:15} and Luong et. al. \shortcite{MAC:13} had used the same training corpus for their models. The corpus contains about 1 billion tokens. For German and French, we use News Crawl (Articles from 2010) released as a part of ACL 2014 Ninth Workshop on Statistical Machine Translation (de-SG and fr-SG respectively). For Urdu, we use the untagged corpus released by Jawaid et. al. ~\shortcite{BJ:15}. For Hindi, we use a monolingual corpus containing 31 million tokens, for training.

As a parallel corpus for Urdu, we use the Hindi-Urdu parallel corpus released by Durrani et. al. \shortcite{ND:10}. We have used the Europarl parallel corpus version 7 ~\cite{PK:02} for parallel sentences of English-French and English-German.

We use standard word-similarity datasets for testing. For English, we use Stanford English Rare-Word (RW) dataset~\cite{MAC:13} and the RG65 dataset ~\cite{RG:65}. The Stanford Rare-Word dataset contains comparatively more rare words and morphological complexity than other datasets and is central to our experiments. For German, we use the German RG65 ~\cite{ZG:06} dataset. For French we use the French RG65 ~\cite{JI:11} dataset. In case of Urdu, we are using the word similarity dataset WS-UR-100 ~\cite{MY:17}.

For testing analogical regularities, we have used the MSR word analogy dataset~\cite{MSG2:13b}. It contains 8000 analogy question. This dataset has been used by us for testing our model. The relations portrayed by these questions are morpho-syntactic, and can be categorized according to parts of speech - adjectives, nouns and verbs. Adjective relations include comparative and superlative (good is to best as smart is to smartest). Noun relations include singular and plural, possessive and non-possessive (dog is to dog's as cat is to cat's). Verb relations are tense modifications (work is to worked as accept is to accepted).

For rest of the paper, we have calculated the Spearman $\rho$ (multiplied by 100) between human assigned similarity and cosine similarity of our word embeddings for the word-pairs. 

All the thresholds mentioned have been decided after empirical fine tuning. Even though our experiments were computationally optimized, time and space complexities also played a part in deciding our thresholds.

In order to learn initial representations of the words, we train word embeddings (word2vec) using the parameters described above on the training set. This model is referred to as SG (Skip-gram).

\section{Cross-Lingual Transformation Matrix}


Since we are generating our transformation matrix from parallel word pairs of two different languages, we first need a list of highly frequent word pairs. For generating this list, we have parallel corpus of two different languages - in our case, Hindi and Urdu, and English and French/German. This parallel corpus contains aligned sentences, from which we generate a one-to-one mapping between words. 

First we obtain word alignments using fast align~\cite{CD:13} which gives us many to many word mapping, which we further use to construct our confidence matrix. This confidence matrix is generated using the frequency count of each individual mapping. For each word of the source language in the confidence matrix, we find out the word in the target language that it has been matched with most frequently and its fraction among all matches. After this, we proceed in decreasing order of count of matches while building a one-to-one matching above a threshold of match count and fraction of matches (25 and 0.5 respectively). 

For this word pair list, we have a frequency threshold, which decides the word pairs that will be chosen for generating our transformation matrix (see Table~\ref{font-table1}). For our experiments, this threshold has been set at 500 (to ensure that they are well trained).

\begin{figure}[h!]
  
  \centering
\includegraphics[width=75mm,scale=0.5]{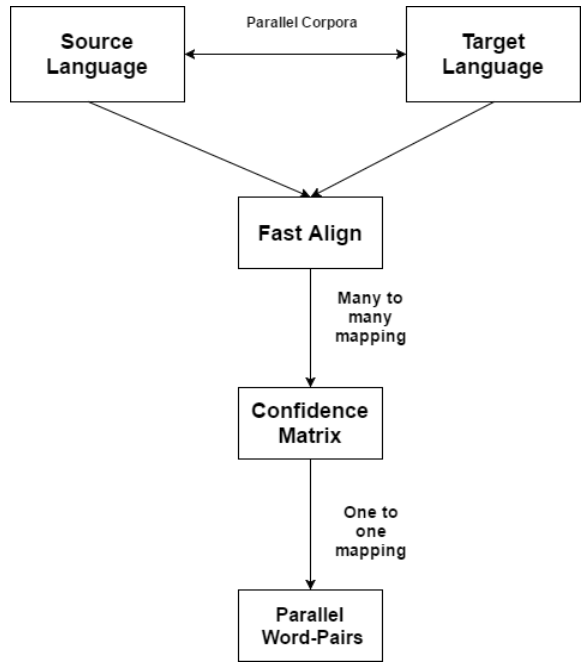}

\caption{Generating Cross-Lingual Word Pairs}
\label{wordpair_workflow}
\end{figure}

\begin{table}[h]
\begin{center}
\caption{\label{font-table1}Example of word pairs for training transformation matrix of English-French from the confidence matrix. ``E. Word" is the word in English, ``F. Word" is the word in French, Count is the number of times ``E. Word" was aligned with ``F. Word" and Fraction is the fraction of the count over all the times ``E. Word" was aligned with any word. }
\begin{tabular}{|l|c|c|c|}
\hline \bf E. Word & \bf F. Word & \bf Count & \bf Fraction \\ \hline
and & et & 1.1M & 0.87 \\
of & de & 1M & 0.51 \\
that & que & 422K & 0.51 \\
we & nous & 319K & 0.65\\
not & pas & 229K & 0.55 \\
We & Nous & 108K & 0.63 \\
Mr & Monsieur & 99K & 0.5\\
- & - & 98K & 0.79 \\
\hline
\end{tabular}
\end{center}
\end{table}

Figure~\ref{wordpair_workflow} gives a high level overview of how we generate highly frequent word pairs from parallel corpus of source and target language in an unsupervised way.

\begin{figure}[h!]
  
  \centering
  
\includegraphics[width=75mm,scale=0.5]{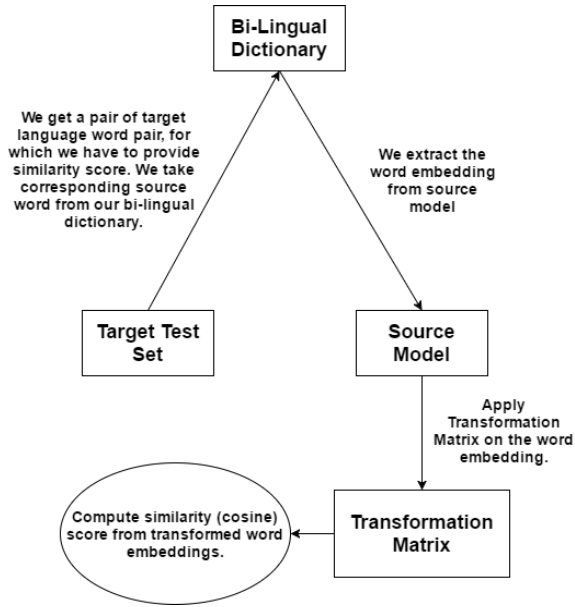}
\caption{Computing similarity scores via transformation matrix.}
\label{transformation}
\end{figure}

Suppose we get ``N" highly frequent word pairs. Dimensions of word embedding of a word in our model is ``D". Using first word of our ``N" chosen word pairs, we create a matrix ``A" of dimensions N*D, where each row is vector representation of the first word. Similarly, we create another matrix B, of similar dimensions as A, using second word of our chosen word pairs.

We now propose that a matrix ``X" (our transformation matrix) will exist such that A*X = B, i.e. X = \inv{A}*B. Our matrix ``X" will be of dimensions ``D*D" and when applied to a word embedding (matrix of dimensions 1*D, it gives a matrix of dimensions 1*D as output), it results in the word embedding of the transformed form of the word. Due to inverse property of a matrix, it accurately remembers the word pairs used for computing it. The matrix also appears to align itself with the word embedding of other words (not used for its training) to transform them according to the patterns that the matrix follows.

\subsection{Transformation between Similar Language Pairs}

We tried our approach on a very similar language pair - ''Hindi-Urdu" with Hindi as the source language and Urdu as the target language. Durrnai et. al. \shortcite{ND:10} observed that 62\% of the Hindi vocabulary are also a part of the Urdu vocabulary after transliteration. The approach smoothly maps embeddings between pairs of linguistically similar or derivative languages, as well as between languages of diverse linguistic properties. Similarity between languages further improves the performance as is evident from results shown in table  ~\ref{font-table7}. 
\begin{table}[h]
\begin{center}
\caption{\label{font-table7} WS-UR-100 is the \textbf{Urdu} version of en-RG-65 and en-WS353 test sets. Note that hi-SG-T denotes the scores of transformed word embeddings of hi-SG. We see that there is a large increase in the scores of the transformed embeddings of hi-SG-T, considerably greater than en-SG-T or de-SG-T over en-SG .}
\begin{tabular}{|l|c|c|c|}
\hline \bf System & \bf WS-UR-100 & \bf Vocab \\ \hline
ur-SG & 34.50 & 130K   \\
hi-SG-T & \textbf{50.08} & -   \\
\hline
\end{tabular}
\end{center}

\end{table}

\subsection{Transformation between Diverse Language Pairs}

\begin{table}[h]
\begin{center}
\caption{\label{font-table4} This table denotes the results of various systems on en-RG-65 test set. SO denotes the scores of Soricut and Och \shortcite{SO:15}. ConceptNet denotes the results of ensemble approach by Speer and Chin \shortcite{RS:16} and en-SG denotes the scores of the model trained by us. This table is for comparison between en-SG and ConceptNet and how this difference is reflected when these models are transformed to other languages. }
\begin{tabular}{|l|c|c|c|}
\hline \bf System & \bf en-RG-65 & \bf Vocab \\ \hline
SO  \shortcite{SO:15} & 75.1 & 1.2M \\
SO w/ Morph \shortcite{SO:15} & 75.1 & 1.2M \\
\hline
en-SG & 74.12 & 2.4M   \\
ConceptNet & \textbf{90.16} & 0.4M   \\
\hline
\end{tabular}
\end{center}

\end{table}

\begin{table}[h]
\begin{center}
\caption{\label{font-table5} fr-RG-65 is the \textbf{French} version of en-RG-65 test set. Note that en-SG-T and ConceptNet-T denote the scores of transformed word embeddings of en-SG and ConceptNet. We can see that en-SG gave a score of 74.12 on en-RG-65 and 68.62 on fr-RG-65 after transformation, whereas for ConceptNet, the scores are 90.16 and 80.13 respectively.}
\begin{tabular}{|l|c|c|c|}
\hline \bf System & \bf fr-RG-65 & \bf Vocab \\ \hline
SO  \shortcite{SO:15} & 63.6 & 1.2M \\
SO w/ Morph \shortcite{SO:15} & 67.3 & 1.2M \\
\hline
fr-SG & 62.48 & 0.5M   \\
en-SG-T & \textbf{68.62} & -   \\
ConceptNet-T & \textbf{80.13} & -   \\
\hline
\end{tabular}
\end{center}

\end{table}

We have used English RG-65 test set as a translation table for French RG-65 and German RG-65 with minor corrections. It is to be noted that the French and German versions of RG-65 were constructed by translating and re-annotating English RG-65. 

We use word similarity task as our evaluation criteria. Given a word pair (French or German), we find its translation in English using English RG-65 test as a translation table. We then find the corresponding word embedding of the word (English) in ConceptNet/SG, and apply our transformation matrix on the word embedding, to generate word embedding for the word in the vector space of target language (French or German) which is further used to generate word similarity score. The scores are shown in Tables ~\ref{font-table4}, ~\ref{font-table5} and ~\ref{font-table6}.

\begin{table}[h]
\begin{center}
\caption{\label{font-table6} de-RG-65 is the \textbf{German} version of en-RG-65 test set. Note that en-SG-T and ConceptNet-T denote the scores of transformed word embeddings of en-SG and ConceptNet. We see that there is a proportional increase in the scores of the transformed embeddings of en-SG-T and ConceptNet-T (as we saw in case of French).}
\begin{tabular}{|l|c|c|c|}
\hline \bf System & \bf de-RG-65 & \bf Vocab \\ \hline
SO  \shortcite{SO:15} & 62.4 & 2.9M \\
SO w/ Morph \shortcite{SO:15} & 64.1 & 2.9M \\
\hline
de-SG & 64.96 & 1.8M   \\
en-SG-T & \textbf{67.3} & -   \\
ConceptNet-T & \textbf{83.17} & -   \\
\hline
\end{tabular}
\end{center}

\end{table}

We see that the approach performs better in the case of Hindi-Urdu as compared to English-German and English-French, owing to similarity between the two languages. Please note that the comparison is not made between ConceptNet-T and hi-SG-T but between en-SG-T (of both French and German) and hi-SG-T because anything of similar nature to ConceptNet does not exist for Hindi.

\section{Transformation between Different Models of Same Language}

\begin{table}[h]
\begin{center}
\caption{\label{font-table2}Scores on Stanford Rare word test set. GN denotes the scores of embeddings trained in Google-News word embeddings. SG denotes the scores of the embeddings trained by us. GN+SG denotes the system in which we import the embeddings of any word missing in GN from SG. }
\begin{tabular}{|l|c|c|c|}
\hline \bf System & \bf RW & \bf Unseen Words  \\ \hline
GN & 45.27 & 173 \\
SG & 40.08 & 88 \\
GN+SG & 48.56 & 58 \\
\hline
\end{tabular}
\end{center}

\end{table}

\begin{table}[h]
\begin{center}
\caption{\label{font-table3}This table denotes the scores on MSR word analogy test set. GN denotes the word embeddings trained on Google-News corpus. SG denotes the embeddings trained by us. GN+SG denotes the system in which we import the embeddings of the words missing in GN from SG. CosSum and CosMul denotes two different techniques for solving word analogy \cite{OL:01}. Unseen words is the number of words missing from the model out of 16000.}
\begin{tabular}{|l|c|c|c|}
\hline \bf System & \bf CosSum & \bf CosMul & \bf Unseen Words \\ \hline
GN & 0.646 & 0.67 & 2000 \\
SG & 0.484 & 0.533 & 0 \\
GN+SG & \textbf{0.674} & \textbf{0.701} & 0 \\
\hline
\end{tabular}
\end{center}

\end{table}

While evaluating and testing our approach, we realized that it might be possible to port and use multiple models as and when needed, for a single language. We often encounter models trained for different types of data, for different purposes, but for the same language. Our approach enables us to generate word embeddings for words that are missing in one model but are present in another model. This enables us to reduce number of unseen words encountered, and after careful evaluation, we found that this approach indeed helps us. We tested this approach using word similarity and word analogy tasks and it showed significant improvement in results.

\begin{figure}[h!]
  
  \centering
\caption{Porting multiple models of same language.}
\includegraphics[width=75mm,scale=0.5]{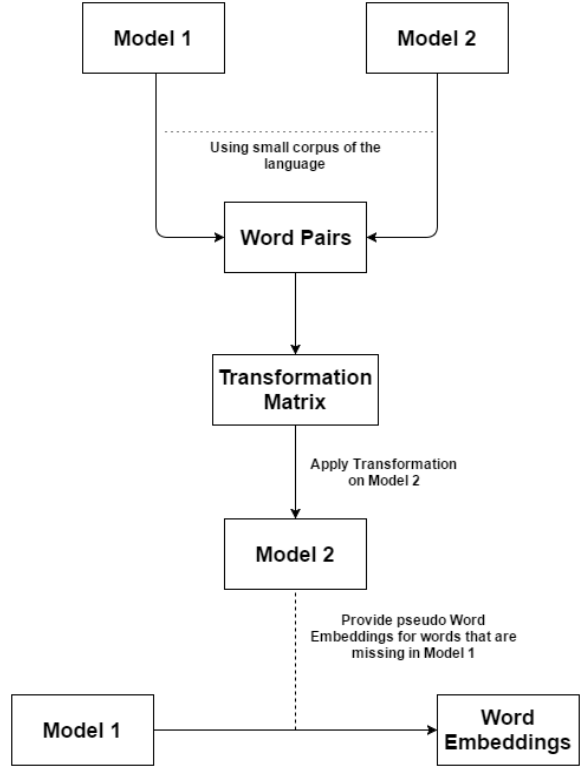}
\label{workflow}
\end{figure}

For creating transformation matrix for two different models of same language, we initially require two different models trained on two different datasets - so that there are certain set of words which are not present in both the models. Then we take a small corpus, for determining the frequent words of the language (above a frequency threshold of 500 and also present in both the models). Using these frequent words, we create our transformation matrix, procedure for which is similar to what we did earlier in section 3. 

After creating this matrix, now whenever we encounter a word which is not present in our first model, we look for the word in our second model, and if found, we apply our transformation matrix to its embedding in our second model. This results in a representation of the word, which proved to be good enough, when we ran word similarity and word analogy tasks on it. 
Figure~\ref{workflow} gives a high level overview of how we try to incorporate different models of the same language via transformation matrix. The scores are shown in tables ~\ref{font-table2} and ~\ref{font-table3}.

\section{Results and Analysis}

The method seems to perform well not only for the same language but also for cross lingual vector space transformation. Cross-lingual vector space transformation results in state of the art results on word similarity test sets of French and German - an increase of 13\% in case of French and 19\% for German (See tables~\ref{font-table5} and~\ref{font-table6}). 

\begin{table}[h]
\begin{center}
\caption{\label{Comparison}Inter-Language comparison of results. ``Increase on SG" denotes the difference between the mentioned systems and the scores of SkipGram embeddings trained on the target languages - fr-SG, de-SG and ur-SG respectively. We are not using ConceptNet-T for this comparison because any equivalent embedding is not available for Hindi. }
\begin{tabular}{|l|c|c|c|}
\hline \bf System & \bf Language Pair & \bf Increase on SG \\ \hline
hi-SG-T & Hindi-Urdu & 15.58 \\
en-SG-T & English-French & 6.14 \\
en-SG-T & English-German & 2.34 \\
\hline
\end{tabular}
\end{center}

\end{table}

We see in table~\ref{Comparison} that for the language pair Hindi-Urdu, there was considerably greater increase than en-SG-T for both English-French and English-German. This may be because Hindi-Urdu are very much more similar than either English-French and English-German.

For transformation within the same language we saw a significant improvement on both word similarity and word analogy test sets for English (See tables~\ref{font-table2} and~\ref{font-table3}).

ConceptNet proved to be state of the art model for English, with significant improvement in quality of word embeddings. However, its not possible to apply techniques like ConceptNet on other languages because of their relative data scarcity. Our approach allows us to overcome this hurdle with encouraging results for languages that are linguistically similar to English. 

\section{Future Work}

While we were able to significantly improve on baseline scores of Urdu and beat previous state of the art systems for German and French, and also improve scores by porting two models to generate word embeddings for missing words, we could further extend this approach by creating transformation matrix for not only two, but for all possible combinations of models available. This way, when ever a particular model is being used for a particular task, we can look for a missing word in other models and use its transformed representation accordingly. Significant improvement in training word embeddings by mapping from a source language to a linguistically similar target language, gives us the hypothesis that similar improvement can be achieved for training between different dialects of the same language and we would like to test the approach further on such pairs.

What we could also do is choosing best available word representation of a word in two or more models. For example, even though we have a word present in our model, but its representation is not reliable because of its low frequency in the corpus it was trained on or we are able to detect somehow that it is not well trained. We could then use a more reliable word embedding from other models, transform it for our model, and then use it. We would still need to run various evaluation tasks on this approach to see its impact and usage in future. We will also try to see if it can be modified to retain its character in one model, and import characteristics from its other word representations in other models, which are relatively more reliable. We could define a heuristic for the same, and evaluate on various different values of parameters in the heuristics.





\end{document}